%% file: PaperForReview.tex
\documentclass[10pt,twocolumn,letterpaper]{article}

\usepackage[pagenumbers]{cvpr} 

\usepackage{graphicx}
\usepackage{amsmath}
\usepackage{amssymb}
\usepackage{booktabs}

\usepackage[acronyms]{glossaries}
\input{acronyms}
\makeglossaries

\usepackage{algorithm}
\usepackage{algpseudocode}

\newtheorem{definition}{Definition}
\newtheorem{theorem}{Theorem}

\usepackage{multirow}

\usepackage{pifont}
\newcommand{\cmark}{\ding{51}}%
\newcommand{\xmark}{\ding{55}}

\makeatletter
\@namedef{ver@everyshi.sty}{}
\makeatother
\usepackage{tikz}
\usepackage{pgfplots}

\usepackage{appendix}

\usepackage[pagebackref,breaklinks,colorlinks]{hyperref}

\usepackage[capitalize]{cleveref}
\crefname{section}{Sec.}{Secs.}
\Crefname{section}{Section}{Sections}
\Crefname{table}{Table}{Tables}
\crefname{table}{Tab.}{Tabs.}

\def\cvprPaperID{11266} 
\def\confName{CVPR}
\def\confYear{2023}

\begin{document}

\title{DPD-fVAE: Synthetic Data Generation Using Federated Variational Autoencoders With Differentially-Private Decoder}

\author{Bjarne Pfitzner and Bert Arnrich\\
Hasso Plattner Institute\\
Ruldolf-Breitscheid-Str. 187, 14482 Potsdam, Germany\\
{\tt\small bjarne.pfitzner@hpi.de} and {\tt\small bert.arnrich@hpi.de}
}
\maketitle

\begin{abstract}
	\Gls{fl} is getting increased attention for processing sensitive, distributed datasets common to domains such as healthcare. 
	Instead of directly training classification models on these datasets, recent works have considered training data generators capable of synthesising a new dataset which is not protected by any privacy restrictions.
	Thus, the synthetic data can be made available to anyone, which enables further evaluation of \acrlong{ml} architectures and research questions off-site.
	As an additional layer of privacy-preservation, differential privacy can be introduced into the training process.
	We propose DPD-fVAE, a federated Variational Autoencoder with Differentially-Private Decoder, to synthesise a new, labelled dataset for subsequent machine learning tasks.
	By synchronising only the decoder component with \gls{fl}, we can reduce the privacy cost per epoch and thus enable better data generators.
	In our evaluation on MNIST, Fashion-MNIST and CelebA, we show the benefits of DPD-fVAE and report competitive performance to related work in terms of Fr\'echet Inception Distance and accuracy of classifiers trained on the synthesised dataset.
\end{abstract}

\glsreset{fl}

\section{Introduction}
\label{sec:introduction}
\input{sections/1_introduction}

\section{Preliminaries}
\label{sec:background}
\input{sections/2_background}

\section{Related Work}
\label{sec:related_work}
\input{sections/2_related_work}

\section{Federated VAEs with Differentiall-Private Decoder (DPD-fVAE)}
\label{sec:dpdvae}
\input{sections/3_methods}

\section{Experiments}
\label{sec:experiments}
\input{sections/4_experiments}

\input{sections/5_discussion}

\section{Conclusion}
\label{sec:conclusion}
\input{sections/7_conclusion}

{\small
\bibliographystyle{ieee_fullname}
\bibliography{PaperForReview}
}

\newpage
\appendix
\appendixpage
\input{supplementary_material}

\end{document}

%% file: acronyms.tex
\newacronym{ml}{ML}{machine learning}
\newacronym{vae}{VAE}{variational autoencoder}
\newacronym{cvae}{cVAE}{conditional variational autoencoder}
\newacronym{sivae}{S-IntroVAE}{soft introspective variational autoenceder}
\newacronym{ae}{AE}{autoencoder}
\newacronym{gan}{GAN}{generative adversarial net}
\newacronym{mlp}{MLP}{multi-layer perceptron}
\newacronym{cnn}{CNN}{convolutional neural network}
\newacronym{pate}{PATE}{Private Aggregation of Teacher Ensembles}

\newacronym{fl}{FL}{federated learning}

\newacronym{dp}{DP}{differential privacy}
\newacronym{rdp}{RDP}{R\'enyi differential privacy}
\newacronym{cdp}{CDP}{central differential privacy}
\newacronym{ldp}{LDP}{local differential privacy}

\newacronym{iid}{i.i.d.}{independent and identically distributed}

\newacronym{fvae}{fVAE}{federated Variational Autoencoder}
\newacronym{dpdfvae}{DPD-fVAE}{federated Variational Autoencoder with Differentially-Private Decoder}

\newacronym{kld}{KLD}{Kullback-Leibler divergence}
\newacronym{sgd}{SGD}{stochastic gradient descent}
\newacronym{adam}{Adam}{adaptive moment estimation}

\newacronym{fid}{FID}{Fr\'echet Inception Distance}
\newacronym{is}{IS}{Inception Score}
\newacronym{1nn}{1-NN}{1-nearest-neighbour}

\newacronym{gdpr}{GDPR}{General Data Protection Regulation}
\newacronym{hipaa}{HIPAA}{Health Insurance Portability and Accountability Act}

\newacronym{cxr}{CXR}{Chest X-Ray}

%% file: sections/1_introduction.tex
The recent advances in \gls{ml} resulted in an ever-growing number of powerful model architectures to gain knowledge from large-scale datasets.
The complexity of many of these (deep) models generally requires a considerable amount of data.
While sufficiently large datasets for \gls{ml} are often theoretically available, they are in practice widely distributed and subject to privacy regulations, such as Europe's \gls{gdpr} or the American \gls{hipaa}. 
An example of an \gls{ml} research field suffering from these restrictions is the healthcare domain, where patient data is particularly hard to access for non-clinical researchers.
Moreover, analysing data from different sources except just one is often vital for the validation of \gls{ml} models to prove the applicability to different data distributions occurring in real life.

\Gls{fl} has been proposed as a means to train \gls{ml} models on distributed datasets~\cite{mcmahan2017communication}. 
It is based on an exchange and averaging of model parameters or parameter updates instead of the actual data. 
That allows training participants, such as hospitals, to contribute to \gls{ml} research with their data, without having to transfer it off-site.

However, it has previously been shown that \gls{fl} alone is not sufficient to guarantee the privacy of sensitive data~\cite{melis2018ExploitingUnintended,hitaj2017DeepModels,wang2019beyond,ziegler2022DefendingAgainst}.
By inspecting the model weights or model updates, adversaries can infer information about the underlying data or even reconstruct it. 
In order to provide training participants with formal guarantees regarding their data's privacy, many papers adopted \gls{dp} along with \gls{fl}. 
It describes the introduction of noise into the model, which hides the impact of clients and their individual data on the trained model. 
In the context of distributed model training, algorithms can either use \gls{cdp}, which protects the data on a user level~\cite{mcmahan2018LearningDifferentially,geyer2018differentially} and hides any clients' participation in the training process, or \gls{ldp}, where every individual data point is protected~\cite{abadi2016deep}.

While \gls{fl} of classification models has been explored extensively in the literature, training of generative models is not as widely researched so far. 
Setting up an \gls{fl} system in real-life can be hard due to a large organisational and legal overhead, as well as necessary data preprocessing steps that allow the same model to be trained by each participant.
It is therefore in many cases beneficial to make use of this lengthy and complicated setup phase to train a data generator instead of a classifier.
The synthetic data can subsequently be used to analyse several different \gls{ml} model architectures, whereas using \gls{fl} for training a classifier directly may only allow one architecture to be evaluated.
Moreover, a data generator can be distributed to many researchers that may have different research questions about the dataset, which have not been considered before.
Finally, conditional data generators can augment and balance out imbalanced real datasets with class or property biases that are common in many domains.

In this work, we focus on training \glspl{vae} on image datasets in an \gls{fl} fashion. 
\Glspl{vae} are capable of generating new, unseen data by learning a continuous, regular latent space representation of the training data and sampling from it~\cite{kingma2013auto}. 
Synthesis of samples from specific classes is made possible by including the original class labels as a conditional into the model. 

Using \gls{dp} for \gls{fl} has been shown to be hard, depending on the number of clients and amount of data, since tight privacy bounds require a lot of added noise~\cite{augenstein2020GenerativeModels,bagdasaryan2019differential}.
We propose a way of reducing the required noise by only training the decoder component of the \gls{vae} using \gls{dp}. 
This affects the $L_2$-norms of client updates, leading to a reduced noise requirement with the same privacy spending.

Our contributions are the following: 
(i) We propose an altered \gls{fl} approach for \glspl{vae}, named \gls{dpdfvae}, which keeps the encoder component private and user-specific while synchronising only the decoder with \gls{dp}. 
(ii) We motivate our method by discussing the interaction of \gls{dp} hyperparameters and their impact on added noise and model performance.
(ii) We evaluate the effectiveness of our approach for non-private \gls{fl}, as well as \gls{fl} with central and local \gls{dp}.

The remainder of this paper is structured as follows.
In \cref{sec:background}, we introduce \glspl{vae}, \gls{fl} and \gls{dp} and \cref{sec:related_work} discusses related work. 
We explain our proposed \gls{dpdfvae} algorithm in \cref{sec:dpdvae}. 
The implementation details, our experimental evaluation and the results are illustrated and discussed in \cref{sec:experiments}. 
Finally, \cref{sec:conclusion} concludes this paper and highlights future work.

%% file: sections/2_background.tex
\subsection{Variational Autoencoders} 
\Glspl{vae}~\cite{kingma2013auto} are neural network-based generative models consisting of two components: an encoder $E$, responsible for encoding an input into some latent space distribution, and a decoder $D$ which reconstructs the input based on the latent information~\cite{goodfellow2016deep}. Conditional \glspl{vae} additionally take label information into account when encoding and decoding an input.

The encoder provides an approximate posterior distribution of the latent variable $z$, $q_{\theta_E}(z|\boldsymbol{x},y)$, while the decoder aims to reconstruct the original data $(\boldsymbol{x}, y)$ from the sampled latent space using variational inference, thus maximising the data log-likelihood $\log p_{\theta_D}(\boldsymbol{x}|y)$.
As a proxy for optimising the log-likelihood, \glspl{vae} use the Evidence Lower Bound (ELBO), which includes the \gls{kld} between the approximate posterior and the prior $p(z|y)$ of the latent variable.

In our experiments, we use the recently proposed $\beta$-\gls{vae}~\cite{higgins2016beta}, which introduces a weight parameter into the loss function to adjust the impact of the \gls{kld} term. The loss $\mathcal{L}_\beta$ consists of the reconstruction loss $\mathcal{L}_{RE}$, which is optimised during training, and the KL-loss $\mathcal{L}_{KL}$, which is used as a regularisation term:
\begin{align} \label{eq:vae_loss}
	\mathcal{L}_\beta &= \mathcal{L}_{RE} - \beta\mathcal{L}_{KL}, \qquad\qquad\text{with}\\
	\mathcal{L}_{RE} &= \mathbb{E}_{q_{\theta_E}}\log p_{\theta_D}(\boldsymbol{x}|z,y) \\
	\mathcal{L}_{KL} &= \text{KL}(q_{\theta_E}(z|\boldsymbol{x},y)||p(z|y))
\end{align}




\subsection{Federated Learning}
\label{subsec:fl}
\gls{fl} was recently introduced as a way to train \gls{ml} models on distributed datasets without requiring direct data access~\cite{mcmahan2017communication}. 
Instead, data owners are responsible for training the model on their data and sending model updates, while a central parameter server aggregates these updates from the participants and applies them to the global model. 


In detail, an \gls{ml} model is trained over a number of global rounds, in which the following three steps are repeated: 
First, the server selects a subset of clients to participate in the current round and sends them the current model weights stored by the server. 
Second, the chosen clients train the model on their local datasets for a number of epochs and send the weight differences between their own and the global model back to the server. 
Finally, the server aggregates the received weight updates and applies the update to the global model. 
A common aggregation strategy is performing a weighted average over the updates, considering the amount of local data each client possesses. 
The three steps are repeated until convergence of for $T$ rounds.

\subsection{Differential Privacy}
\Gls{dp} was previously used mainly in the database domain to quantify the privacy leakage from repeated queries on the same dataset. 
The goal of \gls{dp} is to add sufficient noise to the queries such that there is plausible deniability that any one data sample is included in the database and contributed to the query results. 
Formally, $(\epsilon, \delta)$-\gls{dp} 
is defined as follows:
\begin{definition}[$(\epsilon, \delta)$-\gls{dp} \cite{dwork2014algorithmic}]
    A randomised mechanism $\mathcal{M}: \mathcal{D} \rightarrow \mathcal{R}$ with domain $\mathcal{D}$ and range $\mathcal{R}$ satisfies $(\epsilon, \delta)$-\gls{dp} if for any two adjacent inputs $d, d^\prime \in \mathcal{D}$ and for any subset of outputs $\mathcal{S} \subseteq \mathcal{R}$ it holds that,
    \begin{equation}
        \text{Pr}[\mathcal{M}(d) \in \mathcal{S}] \leq e^\epsilon \text{Pr}[\mathcal{M}(d^\prime) \in \mathcal{S}] + \delta
    \end{equation}
\end{definition}
Adjacent inputs are defined as differing by only the presence/absence of a single sample. Thus, $(\epsilon, \delta)$-\gls{dp} constrains the privacy risk for a mechanism by $\epsilon$, with $\delta$ being the probability that the privacy does not hold.

An important property of \gls{dp} is given by the composition theorem, which enables the computation of the privacy cost of multiple applications of \gls{dp} mechanisms.

\begin{theorem}[Composition of \gls{dp} Mechanisms~\cite{dwork2014algorithmic}]
    Let $\mathcal{M}_i: \mathcal{D} \rightarrow \mathcal{R}_i$ be $(\epsilon_i, \delta_i)$-\gls{dp} for $i \in [k]$. Then if $\mathcal{M}_{[k]}: \mathcal{D} \rightarrow \prod_{i=1}^k\mathcal{R}_i$ is defined to be $\mathcal{M}_{[k]}(x) = (\mathcal{M}_1(x), ..., \mathcal{M}_k(x))$, then $\mathcal{M}_{[k]}$ is $(\sum_{i=1}^k \epsilon_k,\sum_{i=1}^k \theta_k)$-\gls{dp}.
    \label{thm:dp_composition}
\end{theorem}

\subsection{Differential Privacy for Machine Learning}
Abadi et al.~\cite{abadi2016deep} transferred the notion of \gls{dp} to \gls{ml}.
They showed that the privacy cost of each step of optimisation (using \gls{sgd} or another optimiser) can be calculated when performing two additional steps: 
First, the global $L_2$-norm of a batch of gradients is clipped to $S$, which bounds the sensitivity of the set of gradients.
Second, the averaged gradients are noised using Gaussian noise calibrated to the sensitivity with a standard deviation of $zS$, where $z$ is the noise multiplier affecting the privacy cost of the step.  
Due to ~\cref{thm:dp_composition}, the privacy spending of multiple consecutive \gls{dp} optimisation steps can be determined.
The so-called \emph{Moments Accountant}~\cite{abadi2016deep} was introduced as a mechanism to keep track of the privacy spending over multiple epochs. 
Given a privacy budget $\epsilon$, the Moments Accountant can then halt the training, when the budget is about to be depleted.
In practice, we use a \gls{rdp}~\cite{mironov2017renyi} accountant for tracking the privacy spending.
An explanation of \gls{rdp} can be found in \cref{appx:rdp}.

\subsection{Differential Privacy for Federated Learning}
A key reason for using \gls{fl} is the preservation of data privacy. 
However, it was shown that the original \gls{fl} algorithm does not prevent data reconstruction by a malicious client or server~\cite{wang2019beyond,ziegler2022DefendingAgainst}. 
Thus, recent \gls{fl} approaches often use \gls{dp} to provide formal privacy guarantees. 

There are two ways of introducing \gls{dp} into \gls{fl}. 
The first way is \gls{ldp} which corresponds to the application of \gls{dp} optimisation by the data owners, as explained above. 
Thus, all clients are responsible for gradient norm clipping and noising, as well as tracking their own privacy spending.
When the local privacy budget is spent, clients are not available for selection by the server anymore.

The other option for \gls{dp} is \gls{fl}-specific and called \gls{cdp}, or sometimes user-level or client-level privacy~\cite{mcmahan2018LearningDifferentially,geyer2018differentially}.
Here, the clients only clip the $L_2$-norm of their overall model update and send it to the server while the Gaussian noise is added by the server during the aggregation step. 
This requires trust of the clients in the server, since their model updates are not properly secured, and a malicious server could infer information about private data characteristics.

The \gls{dp} guarantees for \gls{ldp} are on a sample level, meaning each data sample is protected, while \gls{cdp} merely protects the data on a client level.

%% file: sections/2_related_work.tex
Synthetic data generation is an active field of research with related works utilising \glspl{vae}~\cite{kingma2013auto}, \glspl{gan}~\cite{goodfellow2020generative}, or other statistical approaches.
Most papers require a central dataset~\cite{chen2018DifferentiallyPrivate,xie2018DifferentiallyPrivate,frigerio2019DifferentiallyPrivate,acs2019DifferentiallyPrivate,torfi2020DifferentiallyPrivate,tantipongpipat2020DifferentiallyPrivate,takahashi2020DifferentiallyPrivate,jiang2022DPVAE,liu2019PPGANPrivacypreserving,beaulieu-jones2019PrivacyPreservingGenerative,ma2020RDPGANRenyiDifferential}, with only a handful using \gls{fl} in their algorithms~\cite{lomurno2021GenerativeFederated,chen2018DifferentiallyPrivate,georgios2021DifferentiallyPrivate,mugunthan2021DPDInfoGANDifferentially,zhang2021FedDPGANFederated,triastcyn2020FederatedGenerative,augenstein2020GenerativeModels,jordon2019PATEGANGenerating,jiang2022PrivacyPreservingHighdimensional,xin2020PrivateFLGAN}.
In the following subchapters, we discuss the federated \gls{gan}- and \gls{vae}-based approaches separately, restricting ourselves to privacy-preserving approaches.

\subsection{Data Generation With \gls{dp}-\glspl{gan}}
The majority of works in the field of differentially-private synthetic data generation use \glspl{gan} as their core model. 
\Glspl{gan} consist of a generator and a discriminator model that are jointly trained using a min-max game with conflicting objectives.
The discriminator aims to differentiate between real and synthetic samples, while the generator tries to synthesise data that fools the discriminator.
The generative capabilities of \glspl{gan} are very high, with recent \gls{gan} architectures being able to synthesise high-quality, realistic image data~\cite{brock2019LargeScale}, however, the training procedure requires a lot of data and suffers from several failure modes where the model does not converge.

Jordon et al.~\cite{jordon2019PATEGANGenerating} transfer the differentially private \gls{pate} framework~\cite{papernot2017SemisupervisedKnowledge,papernot2018ScalablePrivate} to \glspl{gan}.
Augenstein et al.~\cite{augenstein2020GenerativeModels} developed a \gls{cdp} algorithm for \glspl{gan} for the purpose of debugging any issues with client data used for \gls{fl}. Since the generator training step does not require real data, it is performed by the server and only the discriminator is updated locally by each client.
Triastcyn and Faltings~\cite{triastcyn2020FederatedGenerative} describe a similar approach for two clients and additionally propose an alternative \gls{dp} formulation called Differential Average-Case Privacy.
Alternatively, Zhang et al.~\cite{zhang2021FedDPGANFederated} propose a similar approach to ours for \glspl{gan}, where only the generator is synchronised using \gls{fl}. However, the \gls{dp} updates are performed for the discriminator, so that the \gls{dp} is only indirectly introduced into the generator. 
Xin et al.~\cite{xin2020PrivateFLGAN} argue that a parallel training setup is inefficient in terms of data access and \gls{dp} since all selected clients work with an older model in parallel. By instead training the \gls{gan} sequentially, clients can build directly on top of the predecessors' training progress.
Zhang et al.~\cite{zhang2021FedDPGANFederated} transfer \gls{dp} \gls{gan} training to a more complex \gls{gan} structure called InfoGAN.

\subsection{Data Generation With \gls{dp}-\glspl{vae}}
Only three papers and one thesis have investigated the generative capabilities of federated \glspl{vae} to synthesise data.
Chen et al.~\cite{chen2018DifferentiallyPrivate} aim to train \glspl{ae} and \glspl{vae} with strong robustness against three types of attacks. While most of the work is concerned with centralised generator training, one subsection also considers \gls{fl} (but only for the \gls{ae}, not the \gls{vae}). 
Alternatively to traditional \gls{fl} training, Lomurno et al.~\cite{lomurno2021GenerativeFederated} propose locally training data generators with \gls{dp} which are then collected on a generator server as a generator pool. This pool is made available for every contributor to synthesise new data from the joint distribution. While being similar to our work in focussing on the generator/decoder component of the \gls{vae}, it differs substantially by not aggregating the generators using \gls{fl}.
Jiang et al.~\cite{jiang2022PrivacyPreservingHighdimensional} worked on an improved \gls{ldp} algorithm that includes a two-stage dimensionality selection process similar to Liu et al.~\cite{liu2020FedSelFederateda} in order to only aggregate sparse local updates and reduce the privacy cost per step.
All the aforementioned papers don't consider \gls{cdp} in their evaluation, which is likely due to its weaker privacy compared to \gls{ldp} (user-level privacy instead of sample-level privacy).
Georgios~\cite{georgios2021DifferentiallyPrivate} covers both types of \gls{dp} for \gls{fl} in his thesis and trains conditional \glspl{vae} with the capability of generating data for each class. 
As a limitation, his evaluation uses very high privacy budgets for a simulation on a few clients and argues that scaling up the number of clients would reduce the necessary privacy budget.

%% file: sections/3_methods.tex

\input{sections/3_methods/private_encoder}


%% file: sections/3_methods/private_encoder.tex
\begin{figure}
  \centering
  \includegraphics[width=1.0\linewidth]{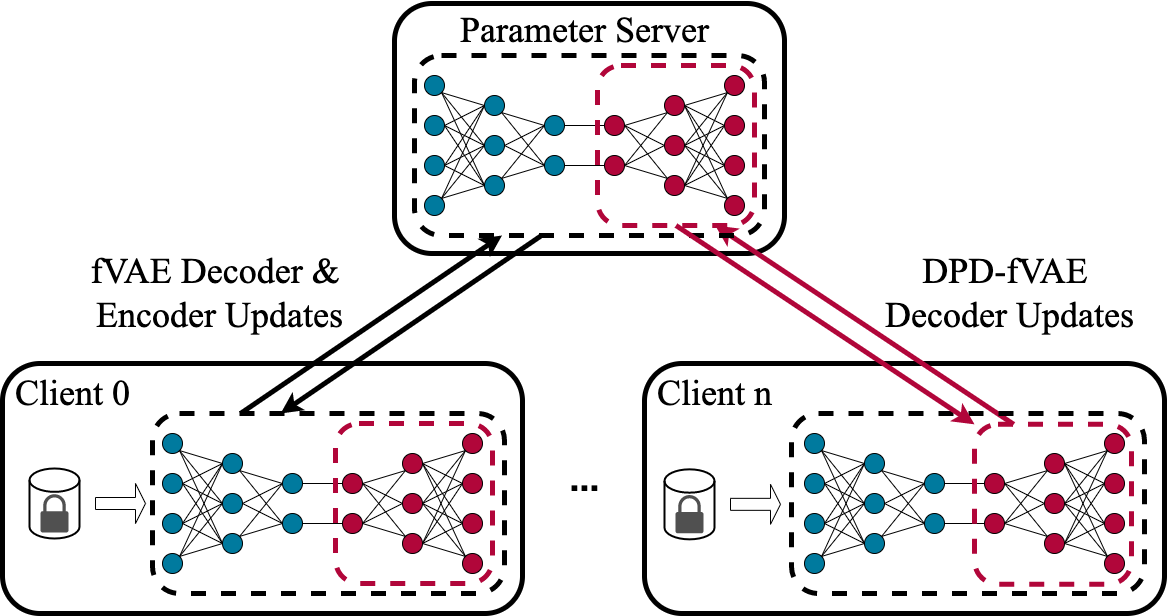}
  \caption{Comparison of vanilla \gls{fvae} and the proposed \gls{dpdfvae}. In the figure, Client 0 is updated using vanilla \gls{fvae}, so the complete \gls{vae} is exchanged between clients and server. Client n stands exemplary for our proposed method, where only the decoder (red) is updated by the clients. The encoders (blue) are only optimised locally and never leave the sites.}
  \label{fig:method_comparison}
\end{figure}

We propose to synchronise only the decoder component of \glspl{vae} with \gls{fl} and \gls{dp}, while the encoders are kept private and personal.
The motivation for this approach is the fact that only the trained decoder is of interest after the training has finished. 
The encoders are not required to synthesise new data.
\cref{fig:method_comparison} illustrates the difference between our proposed \gls{dpdfvae} and vanilla \glspl{fvae}.
\Gls{dpdfvae} can work for \gls{cdp} or \gls{ldp}, which we will term C-\gls{dpdfvae} and L-\gls{dpdfvae}, respectively.
\Cref{alg:pevae_training} shows our training algorithm for \gls{cdp} using \gls{sgd} optimisers for clients and server.
Other optimisers can be used accordingly~\cite{reddi2020adaptive}. 
It is important to note, that stateful optimisers, such as Adam~\cite{kingma2014Adam} or \gls{sgd} with momentum, used for the encoder update step keep their state over multiple global rounds, while the decoder optimiser is reinitialised to account for the changes made during \gls{fl} updates.
The algorithm for \gls{ldp} moves the \gls{dp} clipping, noising and privacy accumulation to the client update (after line 20).
The full algorithm for L-\gls{dpdfvae} can be found in \cref{appx:ldpdfvae_algorithm}.

\begin{algorithm}
\caption{C-\gls{dpdfvae} Training Procedure}
\label{alg:pevae_training}
\begin{algorithmic}[1]
\State {\bfseries Input:} client data $\mathcal{D}_n$ for $N$ clients; maximum global rounds $T$; client sampling probability $q$; local epochs $E$; batch size $B$; local learning rate $\eta$; privacy budget $\epsilon$, privacy risk $\delta$; $L_2$-norm clip $S$; noise multiplier $z$;
\State {\bfseries Server Initialises:} decoder weights $\theta^0$; global privacy accountant $\mathcal{M}((\epsilon, \delta), z, q)$
\State {\bfseries Clients Initialise:} local encoder weights $\Theta_n$ 
   	\vspace{1em}
   	\For{$t=1$ {\bfseries to} $T$}
		\State $\mathcal{C}^t \leftarrow$ randomly select clients with probability $q$
   		\For{$n$ {\bfseries in} $\mathcal{C}^t$ in parallel}
   			\State $\Delta\theta_n^t =$ \textproc{CDP\_Client\_Update}$(n,\theta^{t-1}, E, B)$
   		\EndFor
   		\State $\Delta\theta^t \leftarrow \frac{1}{qN}(\sum_{n \in \mathcal{C}^t} \Delta \theta^t_n + \mathcal{N}(0, z^2 S^2))$
   		\State $\theta^t \leftarrow \theta^t + \Delta\theta^t$
   		\State $\epsilon^t \leftarrow \mathcal{M}\mathtt{.get\_privacy\_spent}(t, z, q)$
   		\If{$\epsilon^t > \epsilon$}
   			\State {\bfseries break}
   		\EndIf
   \EndFor
   \State {\bfseries return} $\theta^t$
   \vspace{1em}
   	\Function{CDP\_Client\_Update}{$n, \theta, E, B$}
   		\State $\theta_n \leftarrow \theta$
   		\For{$e=1$ {\bfseries to} $E$}
   				\For{local batches $b\in\mathcal{D}_n$, with $|b| = B$}
   					\State $\theta_n, \Theta_n \leftarrow SGD(\eta)(\nabla \mathcal{L}(\theta_n, \Theta_n; b))$
   				\EndFor
   			\EndFor
   		\State $\Delta\theta_n \leftarrow \theta - \theta_n$
   		\State $\Delta\theta_n \leftarrow \Delta\theta_n / \text{max}(1, \frac{||\Delta \theta_n||_2}{S})$
   		\State {\bfseries return} $\Delta\theta_n$
   	\EndFunction
\end{algorithmic}
\end{algorithm}

\paragraph{Benefits for \gls{dp}} 
The privacy spending of differentially-private \gls{fl} is determined solely by the choice of the noise multiplier $z$ and, depending on the type of \gls{dp}, the sampling probability $q$ and the number of global rounds $T$ for \gls{cdp}, and the batch sampling probability $\frac{B}{|\mathcal{D}_n|}$ and local optimisation steps $E\times\lceil \frac{B}{|\mathcal{D}_n|}\rceil$ for \gls{ldp}.
Still, the effectiveness of model training can differ immensely depending on the choice of the $L_2$-norm clipping parameter $S$, which additionally affects the standard deviation $\sigma=zS$ of the added Gaussian noise.
Having a large $S$ can reduce the number of clipped parameters and thus retain more information, however, it also requires more noise. 
Too small $S$ values clip too many parameters, hampering good training progress, even if the added noise is smaller.
Thus, it is beneficial to have smaller $L_2$-norms in the weight updates, allowing for less noise with a sufficiently large number of unclipped parameters. 
The factors impacting the magnitude of the $L_2$-norms in \gls{fl} are the following: 
More local training progress in terms of a higher learning rate, an optimiser with momentum, or more local epochs accumulates larger weight updates and thus larger $L_2$-norms.
Additionally, larger networks with more parameters naturally tend to have larger global $L_2$-norms. 
This highlights the benefit of our proposed approach of only synchronising the decoders, which cuts the number of aggregated model parameters approximately in half (depending on the exact \gls{vae} architecture).
Thereby, smaller values for $S$ can be chosen which, in turn, reduces the required noise for the same privacy spending.
Similarly, for \gls{ldp}, we can use a non-\gls{dp} optimiser for updating the encoder parameters, and only clip and noise the decoder's weights.
Firstly, this doesn't introduce any direct noise into the encoder, and secondly, it again lowers the required noise for the decoder's \gls{dp}-optimiser at the same level of privacy guarantee.

As an additional benefit, the smaller number of transmitted parameters also improve the communication efficiency of the algorithm. 
Since this is not the motivation of our work, we leave an in-depth evaluation to future work.

%% file: sections/4_experiments.tex

\subsection{Experiment Setup}
\label{subsec:experiment_setup}
\input{sections/4_experiments/experiment_setup}

\subsection{Non-Private Training}
\label{subsec:non_private_training}
\input{sections/4_experiments/non_private_training}

\subsection{$L_2$-Norm Clipping and Noise Resilience}
\label{subsec:l2_norm_noise_resilience}
\input{sections/4_experiments/norm_noise_resilience}

\subsection{Differentially-Private Training}
\label{subsec:private_training}
\input{sections/4_experiments/private_training}





%% file: sections/4_experiments/experiment_setup.tex
\paragraph{Datasets}
\label{para:datasets}
\input{sections/4_experiments/datasets}

\paragraph{Evaluation Metrics}
\label{para:evaluation_metrics}
\input{sections/4_experiments/evaluation_metrics}

\paragraph{Implementation}
\label{para:implementation}
\input{sections/4_experiments/implementation}

\paragraph{Hyperparameter Optimisation}
\label{para:hyperparameter_optimisation}
\input{sections/4_experiments/hyperparameter_optimisation}

%% file: sections/4_experiments/datasets.tex
To compare with related work, we evaluate our method on three different image datasets: MNIST~\cite{lecun2010mnist}, Fashion-MNIST~\cite{xiao2017FashionMNIST} and CelebA~\cite{liu2015faceattributes}. 
The first two consist of $28\times28$ pixel greyscale images of handwritten digits and clothing articles, respectively, both with ten different classes.
There are $60,000$ images for training and $10,000$ images for testing in both datasets.
We distribute the training data in an \gls{iid} fashion across $500$ and $100$ clients, respectively, resulting in $120$, or $600$ local samples.
We chose two different client numbers to analyse multiple real-world scenarios.

CelebA is a dataset of RGB images of celebrity faces with multiple binary attributes such as facial hair or glasses associated with them.
In our experiments, we use male/female as labels and scale the images down to $32\times32$ pixels.
The distribution of CelebA follows the LEAF framework for benchmarking of \gls{fl}~\cite{caldas2019leaf}, which groups images by celebrity id, excluding those with less than 5 examples.
This results in $9,343$ clients with $19.0\pm7.0$ training samples for a total of $177,457$ training samples.
Approximately $10\%$ of images per celebrity are split off and combined as the central test set with $22,831$ samples.	

%% file: sections/4_experiments/evaluation_metrics.tex
In addition to the subjective visual inspection of synthetic images, all \gls{vae} models are evaluated and compared using two widely used metrics.
The basis for the metric computations are synthetic datasets of $60,000$ images, and we average the scores over five of these datasets.

The first metric is the accuracy of classifiers trained on a synthetic dataset, which shows the utility of the data generator for training \gls{ml} models. 
We use a central test set for evaluating the classifiers, and, like related work, we report results for logistic regression, a \gls{mlp} and a \gls{cnn}~\cite{cao2021DonGenerate}. 
For the specific model architectures, we refer to \cref{appx:model_architectures}.

Secondly, we report the widely-used \gls{fid}~\cite{heusel2017gans}, which determines the similarity of two image datasets.
The \gls{fid} is calculated by evaluating the Fr\'echet Distance~\cite{dumitrescu2004frechet} between the Gaussian distributions of the Inception-v3~\cite{szegedy2016rethinking} network's feature activations.
A lower \gls{fid} score indicates that the synthetic data more closely resembles the original data in terms of similarity and variability.
In our experiments, we compute the \gls{fid} between the synthetic dataset of $60,000$ samples and the complete test set of the datasets.

%% file: sections/4_experiments/implementation.tex

We use TensorFlow 2.5.0\footnote{\url{www.tensorflow.org}} and TensorFlow Federated 0.19.0\footnote{\url{www.tensorflow.org/federated}} for the \gls{fl} code.
For the accumulation of privacy spending, we use the \gls{rdp} accountant supplied by TensorFlow Privacy 0.5.2\footnote{\url{github.com/tensorflow/privacy}}. 
The distributed scenario is simulated by parallelising tasks on a single GPU. 
Our code can be accessed on GitHub\footnote{\url{github.com/BjarnePfitzner/generative_fl_tff}}. 
The \gls{vae} architectures use \glspl{cnn} for encoders and decoders, the model specifics can be found in \cref{appx:model_architectures}.
All experiments were performed on either an NVIDIA A100 40GB, an NVIDIA A40 48GB or an NVIDIA RTX Titan 24GB GPU.

%% file: sections/4_experiments/hyperparameter_optimisation.tex
For each experiment, we performed a hyperparameter optimisation over a predefined discrete search space. 
Where the total number of parameter combinations was below 150, we evaluated all of them in a grid search, and used a random search strategy over 100 trials otherwise.
An exact overview of the evaluated hyperparameter values can be found in the \cref{appx:hyperparams}.

Before doing any \gls{fl} experiments, we evaluated different model architectures, latent dimensions and $\beta$ values using centralised \gls{vae} training. 
Similarly to Lomurno et al.~\cite{lomurno2021GenerativeFederated} and supported by \cite{burgess2018UnderstandingDisentangling}, we searched $\beta$ values between zero and one to emphasise image clarity and found that $\beta=0.01$ performed best. 
These model and loss hyperparameters were then fixed for all following experiments.

For the non-private experiments, we optimised the batch size $B$, the number of local epochs $E$, the client sampling probability $q$, the clients' optimiser and learning rate $opt(\eta)$, and the global momentum $\rho$ (for a constant global \gls{sgd} optimiser with learning rate $1.0$).

For the \gls{dp} experiments, we first defined the \gls{dp} privacy level in terms of $\epsilon$ and $\delta$.
In all experiments, we use $\delta = 10^{-5}$ which is commonly used in related work and sufficiently close to the suggested $\frac{1}{|\mathcal{D}|}$ for our datasets.
For $\epsilon$, we default to $10.0$ but show some results for a low privacy budget of $1.0$.
We trained the models until the privacy budget $\epsilon$ was spent or until a maximum of $1,000$ rounds.
We optimised the two \gls{dp} hyperparameters, the $L_2$-norm clipping threshold $S$ and the noise multiplier $z$, in addition to the aforementioned \gls{fl} hyperparameters. 



%% file: sections/4_experiments/non_private_training.tex
First, we establish a comparable performance in the non-private case of federated \gls{vae} training with only synchronising the decoder, against full \gls{vae} synchronisation. 
Thus, we do not perform any clipping or noising and simply restrict the \gls{fl} synchronisation step to the decoder component of the \gls{vae}.
In \cref{tab:non_private_results}, we show the results for the \gls{cnn} accuracy and \gls{fid}, trained on datasets synthesised by our method (without \gls{dp}) and \gls{fvae}. 
As a baseline, we also trained the \glspl{vae} on a centralised training set. 

\begin{table}
  \centering
  \caption{Results for the non-private variants of \gls{fvae} and \gls{dpdfvae}, compared to a \gls{vae} trained on a centralised dataset.}
  \label{tab:non_private_results}
  \begin{tabular}{llcc}
  	\toprule
    Dataset & Model & FID & \gls{cnn} Acc. ($\%$)\\
    \midrule
    \multirow{3}{*}{MNIST} & Central \gls{vae} & 28.2 & 97.0 \\
    & \gls{fvae} & 30.6 & 96.7 \\
    & \gls{dpdfvae} & 25.3 & 97.3 \\
    \midrule
    \multirow{3}{*}{Fashion-MNIST} & Central \gls{vae} & 56.2 & 82.4 \\
    & \gls{fvae} & 44.6 & 83.3 \\
    & \gls{dpdfvae} & 73.3 & 76.7 \\
    \bottomrule
  \end{tabular}
\end{table}

The performance of our approach is a comparable to vanilla \gls{fvae} and centralised \gls{vae} training, only performing slightly worse for FashionMNIST. 
The hyperparameter optimisation showed, that both federated \gls{vae} trainings are very robust and achieve good scores across many different combinations.
Thus, we conclude, that synchronising only the decoder does not negatively impact the performance of the data generators.

%% file: sections/4_experiments/norm_noise_resilience.tex
\begin{figure}
  \centering
  \begin{subfigure}[b]{0.45\columnwidth}
	  \input{plots/L2norms_small}
    	\vspace{-1em}
		\caption{Smaller Model}
		\label{fig:l2norms_small_model}
	\end{subfigure}
	\hfill
	\begin{subfigure}[b]{0.45\columnwidth}
		\input{plots/L2norms_large}
    	\vspace{-1em}
		\caption{Larger Model}
		\label{fig:l2norms_large_model}
	\end{subfigure}
	\caption{Comparison of the median $L_2$-norms in the first $50$ global training rounds for FashionMNIST. In \cref{fig:l2norms_small_model}, a smaller model was trained ($\approx 185,000$ variables), while the model in \cref{fig:l2norms_large_model} had more parameters ($\approx 3,850,000$ variables) and uses a log-scale. Note, that both larger models and more local epochs correspond to a larger difference in $L_2$-norms.}
  \label{fig:l2norms}
\end{figure}
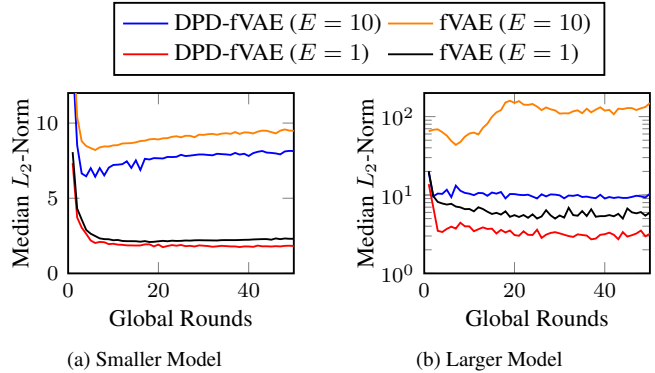

In preliminary experiments for \gls{dp} training, we show the improved resilience of \gls{dpdfvae} against central $L_2$-norm clipping and added Gaussian noise, the two components of \gls{cdp} for \gls{fl}.
It should be evident, that the $L_2$-norms of the client updates are smaller for \gls{dpdfvae} than for \gls{fl} with full synchronisation when equating all hyperparameters, which we illustrate in \cref{fig:l2norms}. 
This effect is increased for more local updates in terms of a higher number of local epochs or a smaller batch size.
The difference is also dependant on the size of the network.

\begin{figure}
  \centering
  \begin{subfigure}{\columnwidth}
		\includegraphics[width=\linewidth]{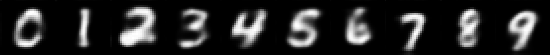}
		\includegraphics[width=\linewidth]{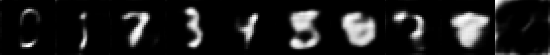}
    \caption{Only $L_2$-clipping at $S=0.2$}
    \label{fig:clip_only}
  \end{subfigure}\\[1ex]
  \begin{subfigure}{\columnwidth}
	\includegraphics[width=\linewidth]{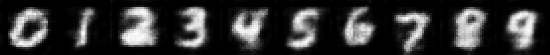}
	\includegraphics[width=\linewidth]{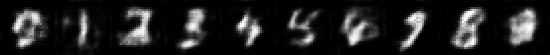}
    \caption{Only added noise with $\sigma=0.05$}
    \label{fig:noise_only}
  \end{subfigure}
  \caption{Synthetic images generated by \gls{dpdfvae} (upper rows) and vanilla \gls{fvae} (lower rows) either using $L_2$-norm clipping \subref{fig:clip_only} or adding Gaussian noise to the global update \subref{fig:noise_only}. \Gls{dpdfvae} outperforms \gls{fvae} in both cases, which together constitute \gls{fl} with \gls{cdp}.}
  \label{fig:clip_noise_resilience}
\end{figure}

Consequently, we can assume a higher resilience against $L_2$-norm clipping for \gls{dpdfvae}. 
We show this in \cref{fig:clip_only}, where the generated samples for a global $L_2$-norm clipping at $S=0.2$ are a lot clearer using \gls{dpdfvae} compared to vanilla \gls{fvae}.


In a second experiment, we investigate the noise resilience of \gls{dpdfvae}. 
By setting different noise levels (in terms of the standard deviation $\sigma$), we compared the resulting synthetic images from both \gls{dpdfvae} and \gls{fvae}.
In \cref{fig:noise_only}, we show the results for a noise level of $\sigma=0.05$ with which \gls{dpdfvae} can still generate comparably clear images, while vanilla \gls{fvae} struggles a lot more with the added noise.

%% file: plots/L2norms_small.tex
\begin{tikzpicture}
    \pgfplotstableread{plots/L2norms_small_data.txt}\data
    \pgfplotsset{
        height=2.4cm,
        width=0.8\textwidth,
        compat=1.3,
        tick label style={font=\footnotesize},
        scale only axis,
    }
    \begin{axis}[
    	x label style={at={(axis description cs:0.5,-0.15)},anchor=north, font=\small},
    	y label style={at={(axis description cs:-0.12,.5)},anchor=south, font=\small},
        xlabel={Global Rounds},
        ylabel={Median $L_2$-Norm},
        xmin=0, xmax=50,
        ymin=0, ymax=12,
        legend columns=2,
        legend style={at={(0.2,1.3)}, anchor=west, font=\small},
        legend cell align={left},
        no markers,
        line width=0.25mm
    ]
        \addplot[blue] table[x=Step,y=DPD-fVAE_E10] {\data};
        \addlegendentry{\gls{dpdfvae} ($E=10$)}
        \addplot[orange] table[x=Step,y=fVAE_E10] {\data};
        \addlegendentry{\gls{fvae} ($E=10$)}
        \addplot[red] table[x=Step,y=DPD-fVAE_E1] {\data};
        \addlegendentry{\gls{dpdfvae} ($E=1$)}	
        \addplot[black] table[x=Step,y=fVAE_E1] {\data};
        \addlegendentry{\gls{fvae} ($E=1$)}
    \end{axis}
\end{tikzpicture}

%% file: plots/L2norms_large.tex

\begin{tikzpicture}
    \pgfplotstableread{plots/L2norms_large_data.csv}\data
    \pgfplotsset{
        height=2.4cm,
        width=0.8\textwidth,
        compat=1.3,
        tick label style={font=\footnotesize},
        scale only axis,
    }
    \begin{axis}[
    	ymode=log,
    	x label style={at={(axis description cs:0.5,-0.15)},anchor=north, font=\small},
    	y label style={at={(axis description cs:-0.17,.5)},anchor=south, font=\small},
        xlabel={Global Rounds},
        ylabel={Median $L_2$-Norm},
        xmin=0, xmax=50,
        ymin=1, ymax=200,
        no markers,
        line width=0.25mm
    ]
        \addplot[blue] table[x=Step,y=DPD-fVAE_E10] {\data};
        \addlegendentry{\gls{dpdfvae} ($E=10$)}
        \addplot[orange] table[x=Step,y=fVAE_E10] {\data};
        \addlegendentry{\gls{fvae} ($E=10$)}
        \addplot[red] table[x=Step,y=DPD-fVAE_E1] {\data};
        \addlegendentry{\gls{dpdfvae} ($E=1$)}	
        \addplot[black] table[x=Step,y=fVAE_E1] {\data};
        \addlegendentry{\gls{fvae} ($E=1$)}
        \legend{};
    \end{axis}
\end{tikzpicture}

%% file: sections/4_experiments/private_training.tex
As the main evaluation of our approach, we investigate the performance of \gls{dpdfvae} for private training of data generators.
We chose to evaluate the \gls{cdp} scenario using MNIST and CelebA and the \gls{ldp} scenario using MNIST and Fashion-MNIST.
This selection was made based on the number of clients and the amount of local data present for each dataset.
For \gls{fl} with \gls{cdp}, the privacy cost per global round is largely impacted by the global sampling probability $q$, with lower $q$s leading to more possible rounds before the privacy budget runs out. 
Since the added Gaussian noise is normalised by the expected number of clients per round $qN$, it is important to have many clients so that $q$ can be chosen to be small, while $qN$ is still large enough.

Similarly for \gls{ldp}, the local privacy cost depends on the local sampling probability $\frac{B}{|\mathcal{D}_n|}$ and the normalisation factor $B$. 
Thus, large local datasets are beneficial in this scenario.

MNIST and CelebA are simulated with 500 and 9343 clients, respectively, favouring \gls{cdp}, while MNIST and Fashion-MNIST possess 120 and 600 local samples, respectively, making them good for \gls{ldp}.

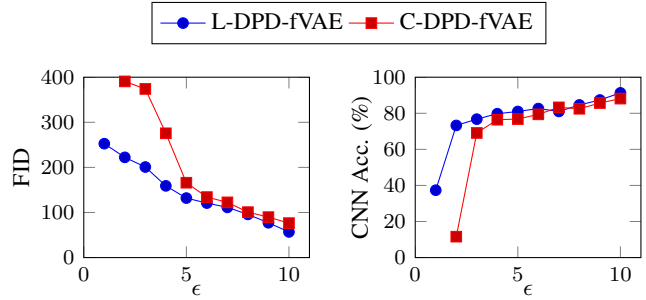
\begin{figure}
  \centering
  \begin{subfigure}[b]{0.45\columnwidth}
	  \input{plots/FID_results_epsilon}
    	\vspace{-1em}
		\label{fig:cdp_mnist_fvae}
	\end{subfigure}
	\hfill
	\begin{subfigure}[b]{0.45\columnwidth}
		\input{plots/ACC_results_epsilon}
    	\vspace{-1em}
		\label{fig:cdp_mnist_pevae}
	\end{subfigure}
    \vspace{-1em}
  \caption{\gls{fid} and \gls{cnn} prediction accuracy for different levels of $\epsilon$ using L-\gls{dpdfvae} and C-\gls{dpdfvae}.}
  \label{fig:private_results_metrics}
\end{figure}

\begin{figure}
    \centering
	\includegraphics[width=\columnwidth]{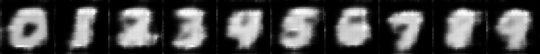}
	\includegraphics[width=\columnwidth]{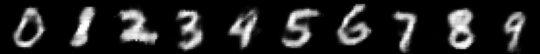}
    \caption{Synthetic MNIST images generated by L-\gls{dpdfvae} with $\epsilon=1$ (upper row) and $\epsilon=10$ (lower row).}
    \label{fig:mnist_ldp_epsilons}
\end{figure}

First, we compare C-\gls{dpdfvae} and L-\gls{dpdfvae} using the MNIST dataset.
\cref{fig:private_results_metrics} shows the data generators' performances for different levels of $\epsilon$.
As expected, smaller privacy budgets result in worse data generators.
For our data distribution with 500 clients and 120 local data points, L-\gls{dpdfvae} still outperforms C-\gls{dpdfvae}, however, this could change for different datasets and distributions.
\cref{fig:mnist_ldp_epsilons} additionally shows images for L-\gls{dpdfvae} with $\epsilon\in{1, 10}$.

\begin{figure}
    \centering
	\includegraphics[width=\columnwidth]{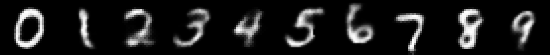}
	\includegraphics[width=\columnwidth]{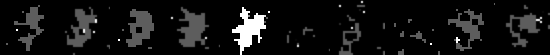}
    \caption{Synthetic MNIST images generated by L-\gls{dpdfvae} (upper row) and \gls{fvae} with \gls{ldp} (lower row), following $(10, 10^{-5})$-\gls{dp}. 
    }
    \label{fig:mnist_ldp_same_hyperparams}
\end{figure}

To illustrate the superiority of \gls{dpdfvae} over vanilla \gls{fvae}, \cref{fig:mnist_ldp_same_hyperparams} shows images generated based on MNIST using the same hyperparameters for both approaches and \gls{ldp}.
It shows that \gls{fvae} fails to converge with the same level of $L_2$-norm clipping, whereas \gls{dpdfvae} can generate high-quality images.

\begin{figure}
  \centering
	\begin{subfigure}{\columnwidth}
		\includegraphics[width=\linewidth]{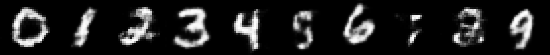}
		\includegraphics[width=\linewidth]{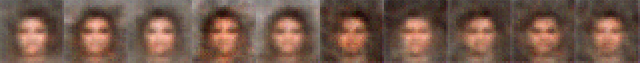}
		\caption{C-\gls{dpdfvae}}
		\label{fig:private_results_cdp}
	\end{subfigure}\\[1ex]
	\begin{subfigure}{\columnwidth}
		\includegraphics[width=\linewidth]{figures/experiments/LDP/sync_d_MNIST}
		\includegraphics[width=\linewidth]{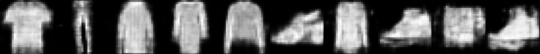}
		\caption{L-\gls{dpdfvae}}
		\label{fig:private_results_ldp}
	\end{subfigure}
	\caption{Synthetic images generated by \gls{dpdfvae} for MNIST, Fashion-MNIST and CelebA using \gls{cdp} \subref{fig:private_results_cdp} and \gls{ldp} \subref{fig:private_results_ldp} with $\epsilon=10$, $\delta=10^{-5}$. 
	}
  \label{fig:private_results_images}
\end{figure}

Finally, in \cref{fig:private_results_images}, we compare synthetic images generated by \gls{dpdfvae} using \gls{cdp} and \gls{ldp} for all three datasets.
The CelebA images are comparatively noisy because of the data distribution provided by TensorFlow Federated, with clients only having very few images. 
This makes even non-private \gls{fl} very difficult.
The synthetic dataset still achieves an \gls{fid} of $261.7$ and a \gls{cnn} accuracy of $69.0$, which is in line with related work.

%% file: plots/FID_results_epsilon.tex

\begin{tikzpicture}
    \pgfplotstableread{plots/epsilons_cdp.txt}\cdp
    \pgfplotstableread{plots/epsilons_ldp.txt}\ldp
    \pgfplotsset{
        height=2.4cm,
        width=0.8\textwidth,
        compat=1.3,
        tick label style={font=\footnotesize},
        scale only axis,
    }
    \begin{axis}[
    	x label style={at={(axis description cs:0.5,-0.1)},anchor=north, font=\small},
    	y label style={at={(axis description cs:-0.2,.5)},anchor=south, font=\small},
        xlabel=$\epsilon$,
        ylabel={FID},
        xmin=0, xmax=11,
        ymin=0, ymax=400,
        legend columns=-1,
        legend style={at={(0.3,1.3)}, anchor=west, font=\small},
        legend cell align={left},
    ]
        \addplot table[x=epsilon,y=fid] {\ldp};
        \addlegendentry{L-\gls{dpdfvae}}
        \addplot table[x=epsilon,y=fid] {\cdp};
        \addlegendentry{C-\gls{dpdfvae}}
    \end{axis}
\end{tikzpicture}

%% file: plots/ACC_results_epsilon.tex

\begin{tikzpicture}
    \pgfplotstableread{plots/epsilons_cdp.txt}\cdp
    \pgfplotstableread{plots/epsilons_ldp.txt}\ldp
    \pgfplotsset{
        height=2.4cm,
        width=0.8\textwidth,
        compat=1.3,
        tick label style={font=\footnotesize},
        scale only axis,
    }
    \begin{axis}[
    	x label style={at={(axis description cs:0.5,-0.1)},anchor=north, font=\small},
    	y label style={at={(axis description cs:-0.15,.5)},anchor=south, font=\small},
        xlabel=$\epsilon$,
        ylabel={CNN Acc. (\%)},
        xmin=0, xmax=11,
        ymin=0, ymax=100,
    ]
        \addplot table[x=epsilon,y=acc] {\ldp};
        \addlegendentry{\gls{ldp} \gls{dpdfvae}}
        \addplot table[x=epsilon,y=acc] {\cdp};
        \addlegendentry{\gls{cdp} \gls{dpdfvae}}
        \legend{};
    \end{axis}
\end{tikzpicture}

%% file: sections/5_discussion.tex
\begin{table*}
    \caption{Results of the evaluation of ($10, 10^{-5}$)-\gls{dp} MNIST and Fashion-MNIST models. The best results are indicated in boldface. All related work results are taken from~\cite{cao2021DonGenerate}, except for DP$^2$-VAE~\cite{jiang2022DPVAE}, who used a slightly different \gls{cnn} implementation, making their accuracy not completely comparable.}
    \label{tab:sota_comparison}
    \begin{center}
    \begin{tabular}{lcccccccccc}
    	\toprule
        \multirow{3}{*}{Method} & \multirow{3}{*}{\gls{fl}} & \multicolumn{4}{c}{MNIST} & & \multicolumn{4}{c}{Fashion-MNIST}\\
        \cmidrule{3-6} \cmidrule{8-11}
        & & \multirow{2}{*}{\gls{fid}} & \multicolumn{3}{c}{Acc. ($\%$)} & & \multirow{2}{*}{\gls{fid}} & \multicolumn{3}{c}{Acc. ($\%$)} \\
        & & & Log. Reg. & MLP & CNN & & & Log. Reg. & MLP & CNN \\
        \midrule
        Real Data & - & 1.1 & 92.6 & 97.5 & 98.9 & & 1.6 & 84.4 & 88.3 & 91.8 \\
        \midrule
        G-PATE~\cite{long2021GPATEScalable} & \xmark & 177.2 & 26 & 25 & 51 & & 205.8 & 42 & 30 & 50\\
        DP-CGAN~\cite{torkzadehmahani2020DPCGANDifferentially} & \xmark & 179.2 & 60 & 60 & 63 & & 243.8 & 51 & 50 & 46 \\
        DP-MERF~\cite{harder2021DPMERFDifferentially} & \xmark & 116.3 & 79.4 & 78.3 & 82.1 & & 132.6 & 75.5 & 74.5 & 75.4\\
        GS-WGAN~\cite{chen2021GSWGANGradientSanitized} & \xmark & 61.3 & 79 & 79 & 80 & & 131.3 & 68 & 65 & 65 \\
        DP-Sinkhorn~\cite{cao2021DonGenerate} & \xmark & \textbf{48.4} & 82.8 & 82.7 & 83.2 & & 128.3 & 75.1 & 74.6 & 71.1\\
        DP$^2$-VAE~\cite{jiang2022DPVAE} & \xmark & 67.6 & \textbf{85.1} & \textbf{87.0} & (90.2) & & 161.6 & \textbf{77.8} & \textbf{76.0} & (76.1) \\
        \midrule
        L-\gls{dpdfvae} & \cmark & 56.9 & 81.4 & 81.8 & \textbf{91.3} & & \textbf{84.4} & 75.0 & 75.4 & \textbf{80.0} \\
        C-\gls{dpdfvae} & \cmark & 76.4 & 79.4 & 81.1 & 88.1 & & N/A & N/A & N/A & N/A\\
        \bottomrule
    \end{tabular}
	\end{center}
\end{table*}

\subsection{Comparison with State-Of-The-Art}
\begin{figure}
    \begin{minipage}{0.23\columnwidth}
		\small
		G-PATE \\[0.9em]
		DP-CGAN \\[0.9em]
		DP-MERF \\[0.9em]
		GS-WGAN \\[0.9em]
		DP-Sinkhorn \\[0.9em]
		DP$^2$-VAE \\[2.8em]
		L-DPD-fVAE \\[2.7em]
		C-DPD-fVAE\\
	\end{minipage}%
	\hfill
    \begin{minipage}{0.77\columnwidth}
	\includegraphics[width=\textwidth]{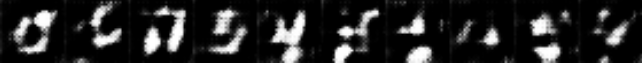}
	\includegraphics[width=\textwidth]{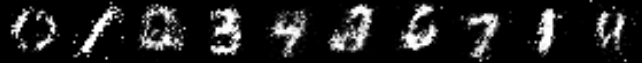}
	\includegraphics[width=\textwidth]{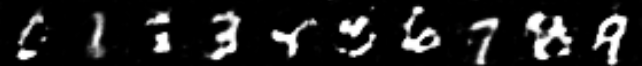}
	\includegraphics[width=\textwidth]{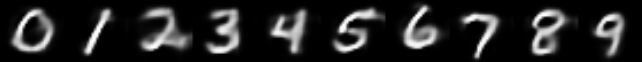}
	\includegraphics[width=\textwidth]{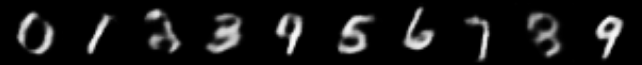}
	\includegraphics[width=\textwidth]{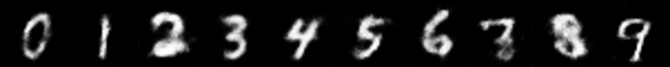}
	\includegraphics[width=\textwidth]{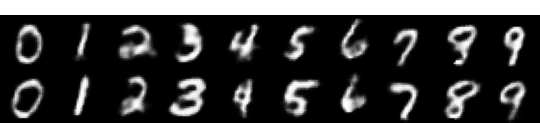}
	\includegraphics[width=\textwidth]{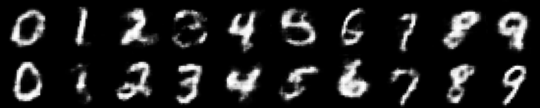}
	\end{minipage}
    \caption{Visual comparison of \gls{dpdfvae} with related work of synthetic MNIST samples under $(10, 10^{-5})$-\gls{dp}.}
    \label{fig:sota_comparison}
\end{figure}

In \cref{tab:sota_comparison}, we show the performance of \gls{dpdfvae} on MNIST and Fashion-MNIST together with results reported by related work, following the same structure as in \cite{jiang2022DPVAE,cao2021DonGenerate}.
Our proposed method achieves very competitive scores compared with related work, even though all comparable papers do not consider \gls{fl}.
In our scenario, the local sampling probability for \gls{ldp} is a lot smaller than for related work, which makes the training task harder (see our motivation in \cref{subsec:private_training}).
\cref{fig:sota_comparison} shows a visual comparison between our reults and previous ones for MNIST. 
For the other datasets, we report more visual results in \cref{appx:additional_results}.

We found that most related works using \gls{fl} for training data generators did not include \gls{dp} in their evaluation and are thus not realistic for privacy-sensitive, real-world data. 
The ones that do consider \gls{dp}~\cite{mugunthan2021DPDInfoGANDifferentially,zhang2021FedDPGANFederated,xin2020PrivateFLGAN} did not publish their code and used different datasets, so we could not compare with them.
\cite{chen2021GSWGANGradientSanitized} include a section on a federated evaluation of their approach, however, they use a very large and unrealistic privacy budget, and their GitHub repository does not include the \gls{fl} code for re-evaluation.

%% file: sections/7_conclusion.tex
We propose \gls{dpdfvae} as a novel federated data generation approach which reduces the negative impact of \gls{dp} on training. 
Our evaluation has shown the effectiveness of our approach on three different image datasets and across various levels of privacy.
Compared with centralised data generation methods, we achieved comparable performance, even if the introduction of \gls{fl} makes the training task more difficult. 

In future work, we aim to investigate the actual resilience of \gls{dpdfvae} against attacks on data privacy.
Additionally, we plan to widen our evaluation by considering a larger set of hyperparameters and also including advanced training techniques such as global adaptive optimisation~\cite{reddi2020adaptive} or per-layer clipping.
We further want to investigate how \gls{dpdfvae} performs using novel \gls{vae} loss functions, such as Soft-IntroVAE~\cite{daniel2021SoftIntroVAEAnalyzing}, or for other dataset modalities such as tabular data or time-series data.
Finally, while related works have investigated federated \glspl{gan} with private discriminators~\cite{fan2020FederatedGenerative,zhang2021FedDPGANFederated}, the benefits for \gls{dp} have not been a focus of these works and would be valuable to analyse. 

%% file: supplementary_material.tex


\section{L-DPD-fVAE}
\label{appx:ldpdfvae_algorithm}
We showed the algorithm for C-\gls{dpdfvae} in the main text. 
The counterpart for \gls{ldp} is shown in \cref{alg:l-dpdfvae_training}.

\section{\Acrlong{rdp}}
\label{appx:rdp}
\input{sections/supplementary/rdp}

\section{Model Architectures}
\label{appx:model_architectures}
\input{sections/supplementary/ldp_algorithm}
\input{sections/supplementary/model_architectures}

\section{Hyperparameters}
\label{appx:hyperparams}
\input{sections/supplementary/hyperparameters}

\section{Additional Results}
\label{appx:additional_results}
\input{sections/supplementary/additional_results}

%% file: sections/supplementary/rdp.tex
\gls{rdp}~\cite{mironov2017renyi} was proposed as an alternative formulation and a natural relaxation of traditional $(\epsilon, \delta)$-\gls{dp}.
It is beneficial for the composition of multiple \gls{ml} optimisation steps with Gaussian noise since the noise scale can be smaller while keeping the same level of privacy compared with $(\epsilon, \delta)$-\gls{dp}.
Formally, \gls{rdp} is defined by:
\begin{definition}[$(\alpha, \epsilon)$-\gls{rdp}~\cite{mironov2017renyi}]
    A randomised mechanism $\mathcal{M}: \mathcal{D} \rightarrow \mathcal{R}$ with domain $\mathcal{D}$ and range $\mathcal{R}$ satisfies $\epsilon$-\gls{rdp} of order $\alpha$ (or $(\alpha, \epsilon)$-\gls{rdp}) if for any two adjacent inputs $d, d^\prime \in \mathcal{D}$ it holds that,
    \begin{align}
       &D_\alpha[\mathcal{M}(d)||\mathcal{M}(d^\prime)] \leq \epsilon,\\
    \text{where}\quad &D_\alpha[P||Q] \triangleq \frac{1}{\alpha-1}\log\mathbb{E}_{x\sim Q}\left(\frac{P(x)}{Q(x)}\right)^\alpha
    \end{align}
\end{definition}

%% file: sections/supplementary/ldp_algorithm.tex
\begin{algorithm}
\caption{L-\gls{dpdfvae} Training Procedure}
\label{alg:l-dpdfvae_training}
\begin{algorithmic}[1]
\State {\bfseries Input:} client data $\mathcal{D}_n$ for $N$ clients; maximum global rounds $T$; client sampling probability $q$; local epochs $E$; batch size $B$; local learning rate $\eta$; privacy budget $\epsilon$, privacy risk $\delta$; $L_2$-norm clip $S$; noise multiplier $z$;
\State {\bfseries Server Initialises:} decoder weights $\theta^0$;
\State {\bfseries Clients Initialise:} local encoder weights $\Theta_n$; local privacy accountant $\mathcal{M}_n((\epsilon, \delta), z, \frac{B}{|\mathcal{D}_n|})$
   	\vspace{1em}
   	\For{$t=1$ {\bfseries to} $T$}
		\State $\mathcal{C}^t \leftarrow$ randomly select clients with probability $q$
   		\For{$n$ {\bfseries in} $\mathcal{C}^t$ in parallel}
   			\State $\Delta\theta_n^t =$ \textproc{LDP\_Client\_Update}$(n,\theta^{t-1}, E, B, \epsilon)$
   		\EndFor
   		\State $\Delta\theta^t \leftarrow \frac{1}{|\mathcal{C}^t|}\sum_{n \in \mathcal{C}^t} \Delta \theta^t_n$
   		\State $\theta^t \leftarrow \theta^t + \Delta\theta^t$
   \EndFor
   \State {\bfseries return} $\theta^t$
   \vspace{1em}
   	\Function{LDP\_Client\_Update}{$n, \theta, E, B, \epsilon$}
   		\State $\theta_n \leftarrow \theta$
   		\For{$e=1$ {\bfseries to} $E$}
   				\For{local batches $b\in\mathcal{D}_n$, with $|b| = B$}
   					\State $\Theta_n \leftarrow SGD(\eta)(\nabla \mathcal{L}(\theta_n, \Theta_n; b))$
   					\State $\theta_n \leftarrow DP\text{-}SGD(\eta)(\nabla \mathcal{L}(\theta_n, \Theta_n; b))$
   					\State $\epsilon_n^t \leftarrow \mathcal{M}_n\mathtt{.get\_privacy\_spent}(t, z, q)$
   					\If{$\epsilon_n^t > \epsilon$}
   						\State remove $n$ from client pool
   						\State {\bfseries return} $\Delta\theta_n$
   					\EndIf
   				\EndFor
   			\EndFor
   		\State $\Delta\theta_n \leftarrow \theta - \theta_n$
   		\State {\bfseries return} $\Delta\theta_n$
   	\EndFunction   
\end{algorithmic}
\end{algorithm}

%% file: sections/supplementary/model_architectures.tex
For all model visualisations we used the Netron tool~\cite{roeder2017Netron}.

\subsection{Variational Autoencoders}
\begin{figure}
	\centering
	\begin{subfigure}{0.57\columnwidth}
  		\includegraphics[width=\textwidth]{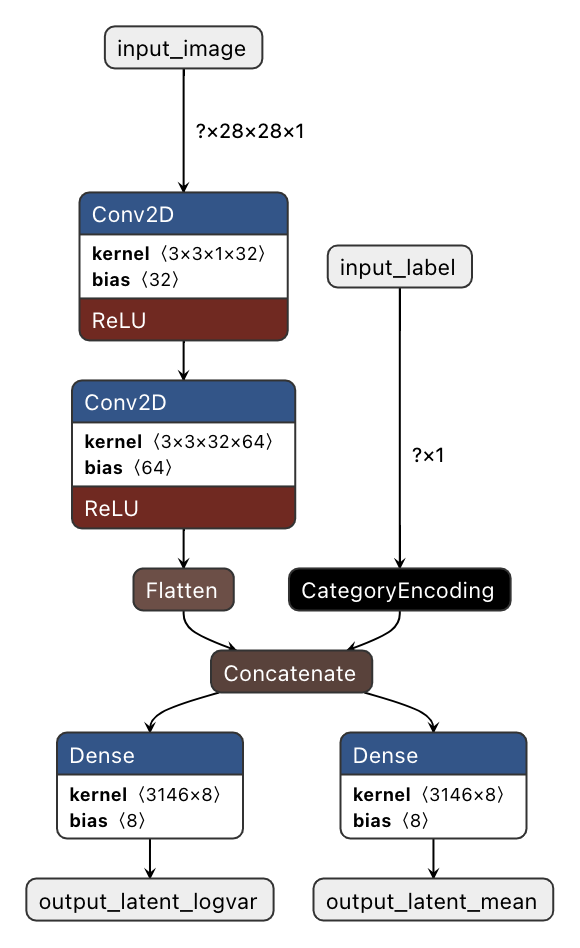}
  		\caption{Encoder}
  		\label{fig:vae_architecture_encoder}
	\end{subfigure}
	\begin{subfigure}{0.41\columnwidth}
  		\includegraphics[width=\textwidth]{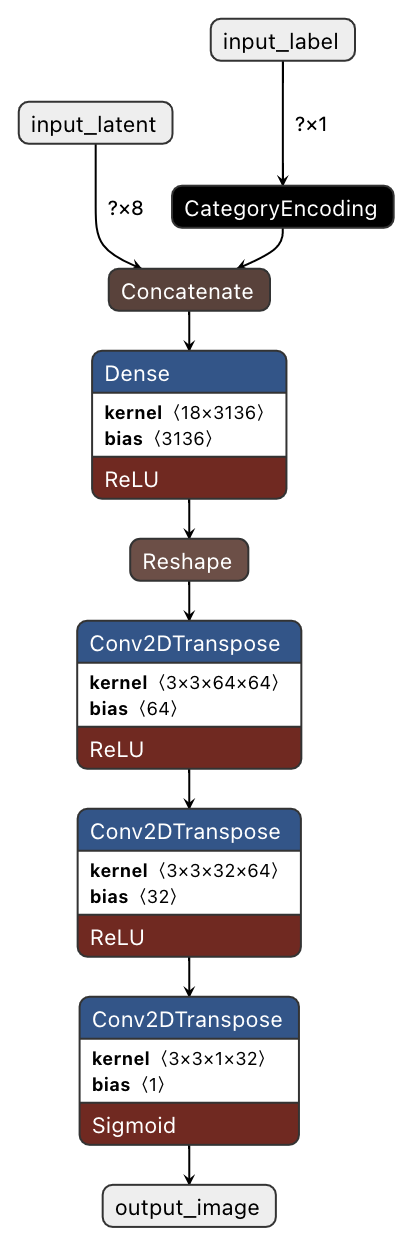}
  		\caption{Decoder}
  		\label{fig:vae_architecture_decoder}
	\end{subfigure}
	\caption{The \gls{vae} model architecture for MNIST and CelebA experiments. All convolutional layers have a stride length of $2$.}
    \label{fig:vae_architecture_small_ccvae}
\end{figure}

\begin{figure}
	\centering
	\begin{subfigure}{0.57\columnwidth}
  		\includegraphics[width=\textwidth]{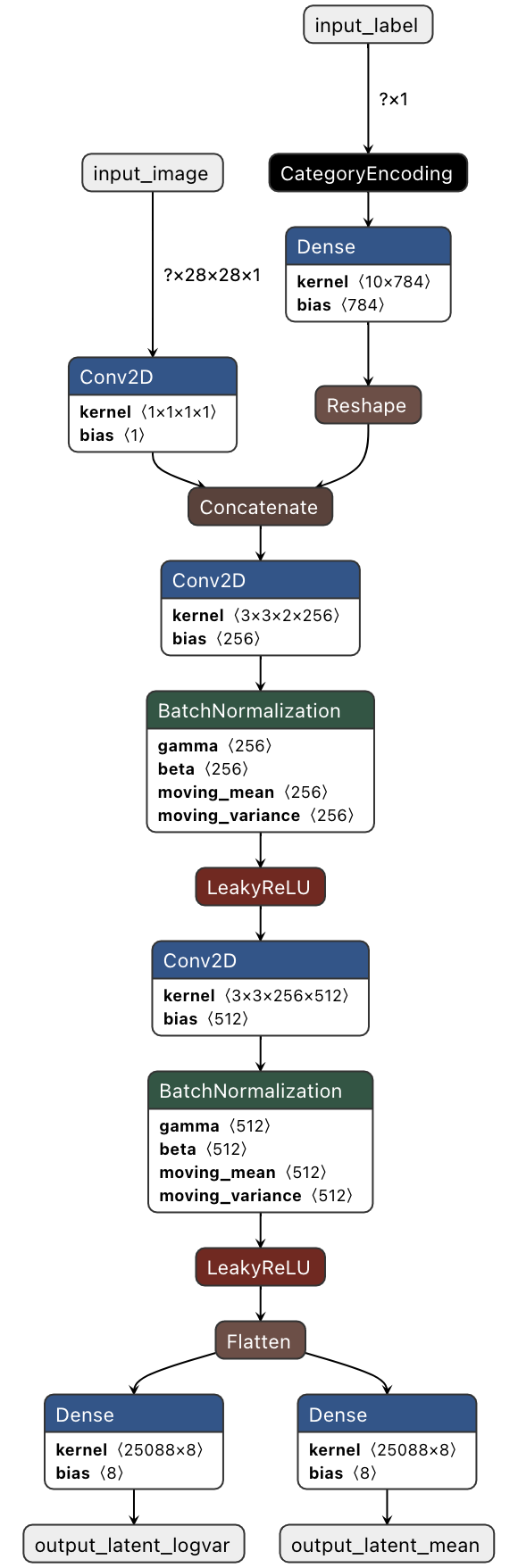}
  		\caption{Encoder}
  		\label{fig:vae_architecture_encoder}
	\end{subfigure}
	\begin{subfigure}{0.41\columnwidth}
  		\includegraphics[width=\textwidth]{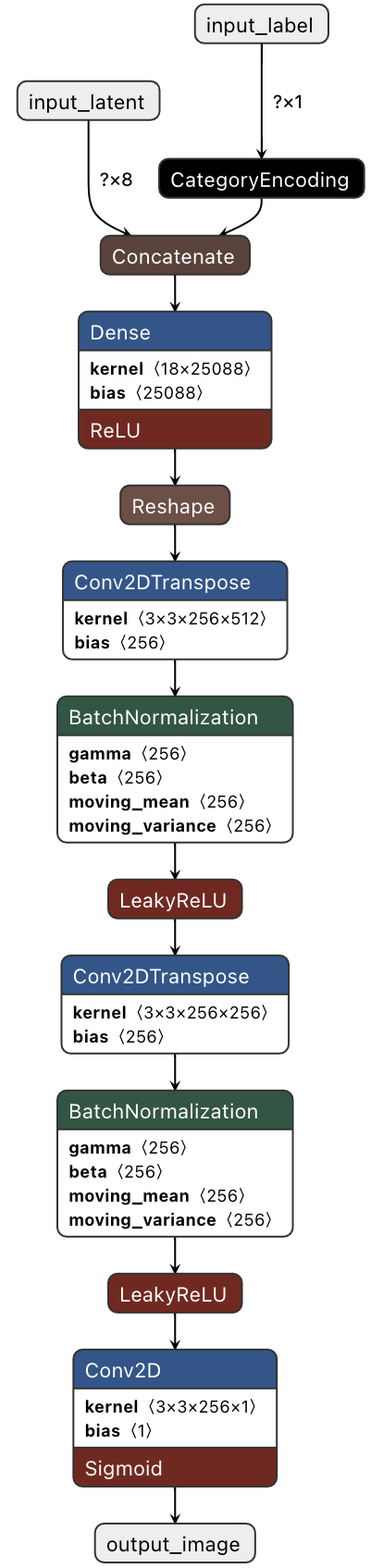}
  		\caption{Decoder}
  		\label{fig:vae_architecture_decoder}
	\end{subfigure}
	\caption{The \gls{vae} model architecture for Fashion-MNIST experiments. The first convolutional layer in the encoder and the last one in the decoder have a stride length of $1$, for all others, the stride length is $2$. The BatchNormalization layers have $\epsilon=10^{-5}$ and a momentum of $0.9$. The LeakyReLU activation uses $\alpha=0.01$.}
    \label{fig:vae_architecture_dp2_ccvae}
\end{figure}

Our \gls{vae} architectures use \glspl{cnn} as their building blocks. 
For the MNIST and CelebA datasets, a smaller architecture shown in \cref{fig:vae_architecture_small_ccvae} performed better, so all presented results are based on that.
For Fashion-MNIST, we used a larger \gls{vae} architecture that was also used by Jiang et al.~\cite{jiang2022DPVAE} and is taken from a public code repository\footnote{\url{https://github.com/AntixK/PyTorch-VAE/blob/master/models/cvae.py}}.
The exact architecture is shown in \cref{fig:vae_architecture_dp2_ccvae}.

\subsection{Evaluation Classifier}
For the evaluation of the usefulness of the synthetic datasets, we use the same three classifiers as Cao et al.~\cite{cao2021DonGenerate}. 

First, we trained a logistic regression (Log. Reg.) with $l_2$-regularisation and $lbfgs$ solver. 

Second, we evaluated a \gls{mlp} with a single hidden layer with $100$ neurons and ReLU activation, and an output layer with one neuron per class and softmax activation. 

Finally, we trained a \gls{cnn} with two convolutional blocks consisting of a 2D-convolution with $32$ and $64$ kernels, respectively, a kernel size of $3$ and stride length $1$, followed by a max pooling layer with stride length $2$, $50\%$ dropout and ReLU activation. 
After these blocks, there is another 2D-convolution with $128$ kernels and otherwise the same setup as the previous ones, however it is directly followed by a ReLU activation (without pooling and dropout).
After that, the output is flattened and fed into another fully-connected layer with $128$ neurons and ReLU activation, $50\%$ dropout and a final fully-connected output layer with one neuron per class and softmax activation.

Both neural networks were trained using the categorical cross-entropy loss and an Adam optimiser with a learning rate of $0.001$. We additionally used a weight decay of $10^{-4}$ for all layers, according to the implementation of \cite{cao2021DonGenerate} found on Github\footnote{\url{https://github.com/nv-tlabs/DP-Sinkhorn_code/blob/main/src/train_mnist_classifier.py}}.

%% file: sections/supplementary/hyperparameters.tex
We investigated a hyperparameter grid with the following options:
\begin{align*}
	\text{local optimiser} &\in \{SGD, Adam\}\\
	\eta &\in \{10^{-2}, 10^{-3}, 10^{-4}, 10^{-5}\}\\
	E &\in \{1, 5, 10\}\\
	\rho &\in \{0.0, 0.5, 0.9, 0.99\}\\
	S &\in \{0.1, 0.2, 0.5, 0.75, 1.0, 1.5, 2.0\}\\
	z &\in \{0.5, 0.6, 0.7, 0.8, 1.0, 1.2, 1.5, 2.0\}
\end{align*}

Since the datasets were set up with different amounts of local data and numbers of clients, the search space for $B$ and $q$ depends on the dataset. 
For MNIST and Fashion-MNIST, we chose $B\in\{10, 20, 30, 60\}$, while for CelebA, we chose $B\in\{8, 16, 32\}$.
The search space for $q$ was $\{0.01, 0.05, 0.1, 0.2\}$, $\{0.05, 0.1, 0.2\}$ and $\{0.005, 0.01, 0.05\}$ for MNIST, Fashion-MNIST and CelebA, respectively.


We present the selected optimal hyperparameters in \cref{tab:hyperparams}.

\begin{table*}
  \centering
  \caption{Optimal hyperparameters used for generating our results. For all experiments a local $Adam$ optimiser performed better than $SGD$. \\ *For the LDP experiment using MNIST, we did not perform hyperparameter optimisation for \gls{fvae}, but used the exact same ones found for \gls{dpdfvae} in order to do a direct comparison.}
  \label{tab:hyperparams}
  \begin{tabular}{llclcccccccccc}
  	\toprule
    Dataset & DP & $(\epsilon, \delta)$ & Model & & $B$ & $\eta$ & & $E$ & $q$ & $\rho$ & & $S$ & $z$\\
    \midrule
    \multirow{9}{*}{MNIST} & \multirow{3}{*}{-} & \multirow{3}{*}{-} & Central \gls{vae} & & 128 & $10^{-3}$ & & - & - & - & & - & - \\
    & & & \gls{fvae} & & 30 & $10^{-2}$ & & 5 & 0.05 & 0.5 & & - & - \\
    & & & \gls{dpdfvae} & & 20 & $10^{-2}$ & & 10 & 0.05 & 0.5 & & - & - \\
    \cmidrule{2-14}
    & \multirow{3}{*}{\gls{cdp}} & \multirow{2}{*}{$(10, 10^{-5})$} & \gls{fvae} & & 10 & $10^{-4}$ & & 10 & 0.2 & 0.0 & & 0.5 & 1.2 \\
    & & & \gls{dpdfvae} & & 20 & $10^{-3}$ & & 10 & 0.2 & 0.5 & & 1.0 & 1.0 \\
    \cmidrule{3-14}
    & & $(1, 10^{-5})$ & \gls{dpdfvae} & & 10 & $10^{-3}$ & & 10 & 0.1 & 0.5 & & 0.5 & 2.0 \\
    \cmidrule{2-14}
    & \multirow{3}{*}{\gls{ldp}} & \multirow{2}{*}{$(10, 10^{-5})$} & \gls{fvae}* & & 20 & $10^{-2}$ & & 5 & 0.01 & 0.5 & & 0.1 & 1.2 \\
    & & & \gls{dpdfvae} & & 20 & $10^{-2}$ & & 5 & 0.01 & 0.5 & & 0.1 & 1.2 \\
    \cmidrule{3-14}
    & & $(1, 10^{-5})$ & \gls{dpdfvae} & & 30 & $10^{-3}$ & & 10 & 0.1 & 0.5 & & 0.1 & 1.8 \\
    \midrule
    \multirow{6}{*}{Fashion-MNIST} & \multirow{3}{*}{-} & \multirow{3}{*}{-} & Central \gls{vae} & & 32 & $10^{-3}$ & & - & - & - & & - & - \\
    & & & \gls{fvae} & & 32 & $10^{-3}$ & & 10 & 0.1 & 0.5 & & - & -\\
    & & & \gls{dpdfvae} & & 32 & $10^{-3}$ & & 10 & 0.1 & 0.5 & & - & -\\
    \cmidrule{2-14}
    & \multirow{3}{*}{\gls{ldp}} & \multirow{2}{*}{$(10, 10^{-5})$} & \gls{fvae} & & 10 & $10^{-4}$ & & 10 & 0.05 & 0.0 & & 0.1 & 0.7 \\
    & & & \gls{dpdfvae} & & 30 & $10^{-4}$ & & 10 & 0.01 & 0.0 & & 2.0 & 1.0 \\
    \cmidrule{3-14}
    & & $(1, 10^{-5})$ & \gls{dpdfvae} & & 10 & $10^{-4}$ & & 1 & 0.01 & 0.0 & & 0.2 & 2.0 \\
    \midrule
    CelebA & \gls{cdp} & $(10, 10^{-5})$ & \gls{dpdfvae} & & 8 & $10^{-2}$ & & 10 & 0.02 & 0.9 & & 0.5 & 0.7\\
    \bottomrule
  \end{tabular}
\end{table*}


%% file: sections/supplementary/additional_results.tex
\begin{figure}
    \begin{minipage}{0.23\columnwidth}
		\small
		G-PATE \\[0.9em]
		DP-CGAN \\[0.9em]
		DP-MERF \\[0.9em]
		GS-WGAN \\[0.9em]
		DP-Sinkhorn \\[0.9em]
		DP$^2$-VAE \\[2em]
		L-DPD-fVAE \\
	\end{minipage}%
	\hfill
    \begin{minipage}{0.77\columnwidth}
	\includegraphics[width=\textwidth]{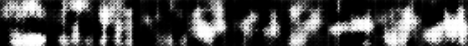}
	\includegraphics[width=\textwidth]{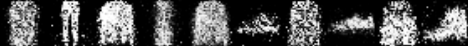}
	\includegraphics[width=\textwidth]{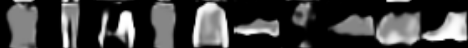}
	\includegraphics[width=\textwidth]{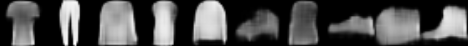}
	\includegraphics[width=\textwidth]{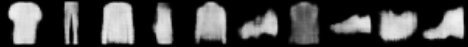}
	\includegraphics[width=\textwidth]{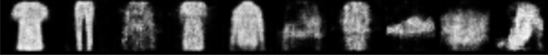}
	\includegraphics[width=\textwidth]{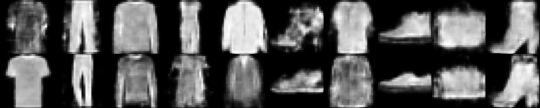}
	\end{minipage}
    \caption{Visual comparison of L-\gls{dpdfvae} with related work of synthetic Fashion-MNIST samples under $(10, 10^{-5})$-\gls{dp}.}
    \label{fig:fmnist_sota_comparison}
\end{figure}
\begin{figure}
	\centering
	\includegraphics[width=0.9\columnwidth]{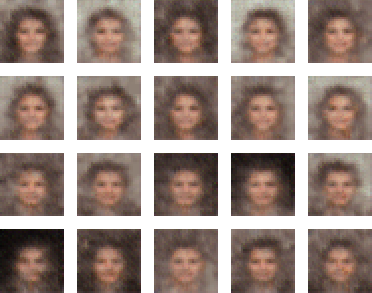}
	\caption{Images generated by C-\gls{dpdfvae} for CelebA with $\epsilon=10.0$. The upper two rows of images correspond to the female class, the bottom ones to the male class. FID $=261.7$.}
	\label{fig:celeba_cdpdfvae}
\end{figure}
\begin{figure*}
	\centering
	\begin{subfigure}{0.45\linewidth}
		\includegraphics[width=\textwidth]{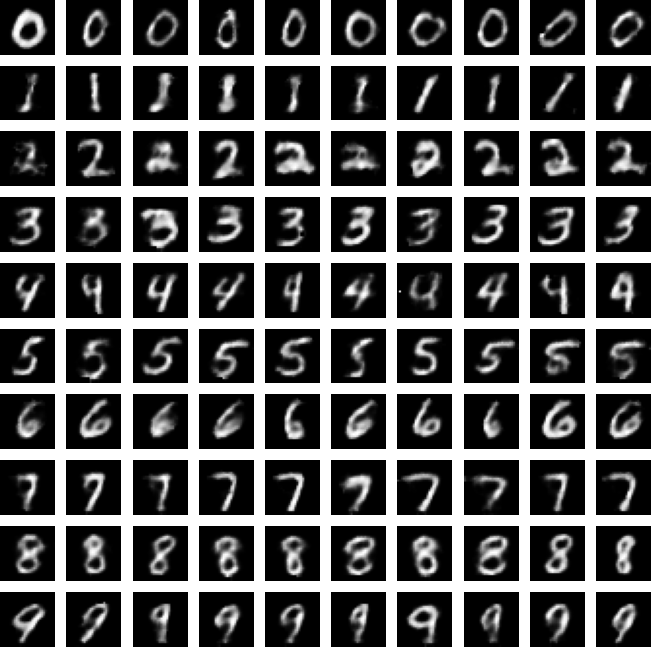}		
		\caption{$\epsilon=10.0$, FID $=56.9$}
	\end{subfigure}
	\hfil
	\begin{subfigure}{0.45\linewidth}
		\includegraphics[width=\textwidth]{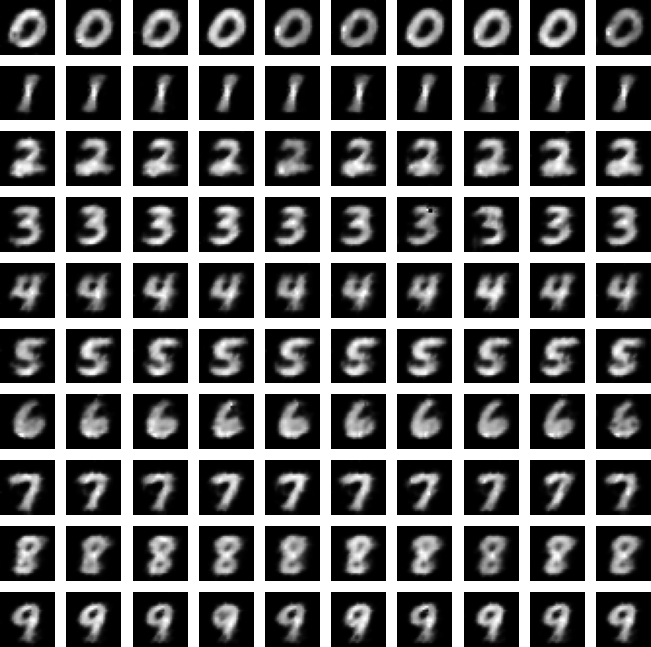}		
		\caption{$\epsilon=1.0$, FID $=151.9$}
	\end{subfigure}
	\caption{Images generated by L-\gls{dpdfvae} for MNIST with two different privacy budgets.}
	\label{fig:mnist_ldpdfvae}
\end{figure*}
\begin{figure*}
	\centering
	\begin{subfigure}{0.45\linewidth}
		\includegraphics[width=\textwidth]{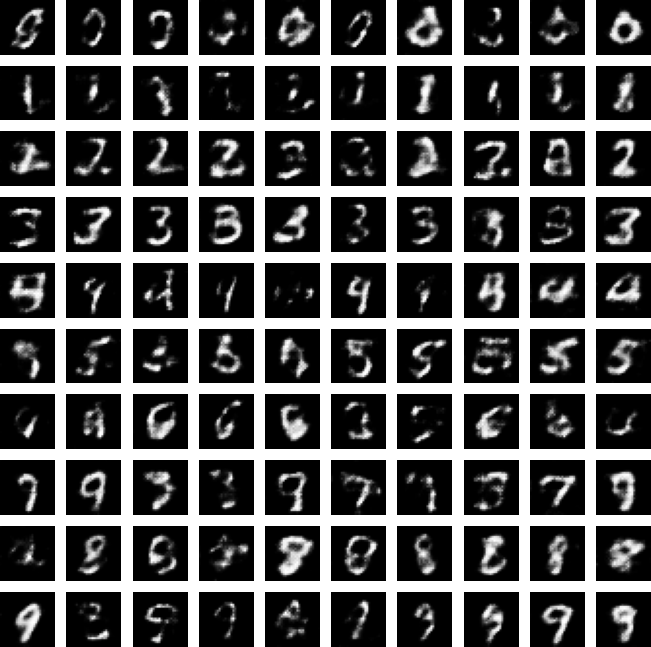}		
		\caption{$\epsilon=10.0$, FID $=76.4$}
	\end{subfigure}
	\hfil
	\begin{subfigure}{0.45\linewidth}
		\includegraphics[width=\textwidth]{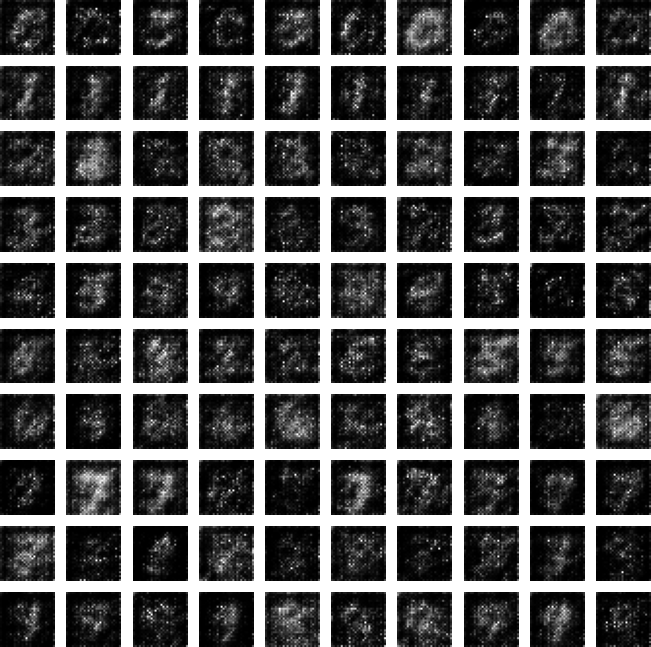}		
		\caption{$\epsilon=1.0$, FID $=295.2$}
	\end{subfigure}
	\caption{Images generated by C-\gls{dpdfvae} for MNIST with two different privacy budgets.}
	\label{fig:mnist_cdpdfvae}
\end{figure*}
\begin{figure*}
	\centering
	\begin{subfigure}{0.45\linewidth}
		\includegraphics[width=\textwidth]{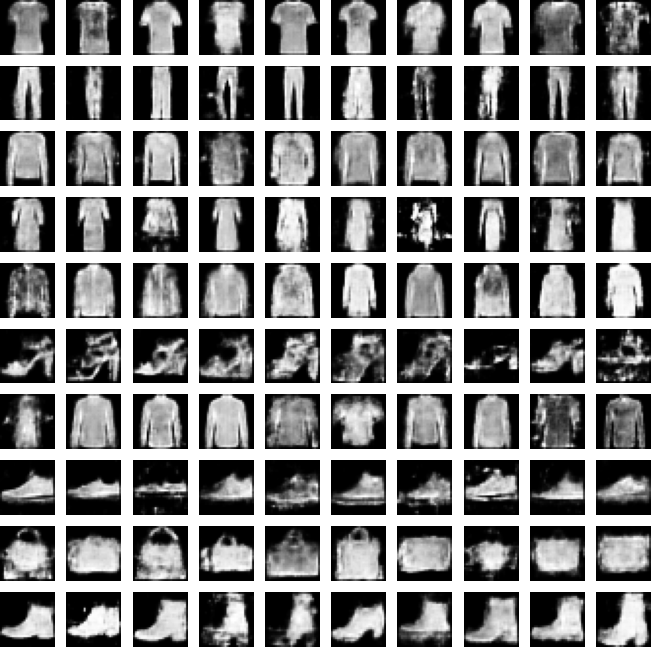}		
		\caption{$\epsilon=10.0$, FID $=84.4$}
	\end{subfigure}
	\hfil
	\begin{subfigure}{0.45\linewidth}
		\includegraphics[width=\textwidth]{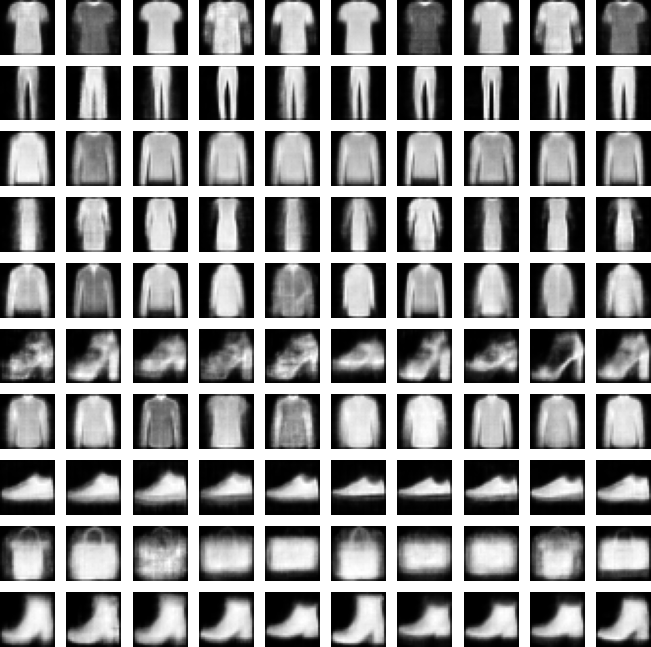}		
		\caption{$\epsilon=1.0$, FID $=149.8$}
	\end{subfigure}
	\caption{Images generated by L-\gls{dpdfvae} for FashionMNIST with two different privacy budgets.}
	\label{fig:fmnist_ldpdfvae}
\end{figure*}

In \cref{fig:fmnist_sota_comparison}, we compare our Fashion-MNIST images with related work.
Moreover, we present additional generated images for all datasets and $\epsilon=10.0$ and $\epsilon=1.0$ (except for CelebA) in \cref{fig:mnist_ldpdfvae,fig:mnist_cdpdfvae,fig:fmnist_ldpdfvae,fig:celeba_cdpdfvae}.

Notably, for training C-\gls{dpdfvae} with $\epsilon=1.0$ on MNIST, we did not find a well-performing model. 
This is likely due to the number of clients being $500$. 
Having more clients would allow more global rounds, because smaller $q$s could be chosen, which reduces the privacy cost per round.

